\documentclass[runningheads]{llncs}

\usepackage{eccv}

\usepackage{eccvabbrv}

\usepackage{graphicx}
\usepackage{booktabs}
\usepackage{colortbl}
\usepackage{multirow}
\newcommand{\tabincell}[2]{\begin{tabular}{@{}#1@{}}#2\end{tabular}} 

\usepackage[accsupp]{axessibility}  

\usepackage{hyperref}

\usepackage{orcidlink}

\begin{document}

\title{Boosting Gaze Object Prediction via Pixel-level Supervision from Vision Foundation Model} 

\titlerunning{Boosting Gaze Object Prediction via Pixel-level Supervision}

\author{Yang Jin\inst{1}\orcidlink{0009-0007-3285-427X} \and
Lei Zhang\inst{2}\orcidlink{0000-0003-1520-9360} \and
Shi Yan\inst{2}\orcidlink{0000-0002-6123-9870} \and
Bin Fan\inst{3}\orcidlink{0000-0002-8028-0166} \and
Binglu Wang\inst{1}$\dagger$\orcidlink{0000-0002-9266-4685}
}

\authorrunning{Y.~Jin et al.}

\institute{Xi’an University of Architecture and Technology, Xi’an, China \\
\email{\{jin91999, wbl921129\}@gmail.com}  
$\quad$ $\dagger$  Corresponding author\\
\and
School of Automation, Northwestern Polytechnical University, Xi’an, China\\
\email{zl\_hnly@163.com, yanshi@mail.nwpu.edu.cn}\\
 \and
National Key Lab of General AI, School of Intelligence Science and Technology, Peking University, Beijing, China\\
\email{binfan@pku.edu.cn}\\
}

\maketitle

\begin{abstract}
Gaze object prediction (GOP) aims to predict the category and location of the object that a human is looking at. Previous methods utilized box-level supervision to identify the object that a person is looking at, but struggled with semantic ambiguity, \ie, a single box may contain several items since objects are close together. The Vision foundation model (VFM) has improved in object segmentation using box prompts, which can reduce confusion by more precisely locating objects, offering advantages for fine-grained prediction of gaze objects. This paper presents a more challenging gaze object segmentation (GOS) task, which involves inferring the pixel-level mask corresponding to the object captured by human gaze behavior. 
In particular, we propose that the pixel-level supervision provided by VFM can be integrated into gaze object prediction to mitigate semantic ambiguity. This leads to our gaze object detection and segmentation framework that enables accurate pixel-level predictions. Different from previous methods that require additional head input or ignore head features, we propose to automatically obtain head features from scene features to ensure the model’s inference efficiency and flexibility in the real world. Moreover, rather than directly fuse features to predict gaze heatmap as in existing methods, which may overlook spatial location and subtle details of the object, we develop a space-to-object gaze regression method to facilitate human-object gaze interaction. Specifically, it first constructs an initial human-object spatial connection, then refines this connection by interacting with semantically clear features in the segmentation branch, ultimately predicting a gaze heatmap for precise localization. Extensive experiments on GOO-Synth and GOO-Real datasets demonstrate the effectiveness of our method. The code will be available at \url{https://github.com/jinyang06/SamGOP}.
\end{abstract}

\section{Introduction}
The objects stared at by humans contain important semantic information, which can reveal a person's behavior and state of mind. Therefore, identifying the objects that people are looking at has a wide range of applications. For example, in medical diagnosis, whether a child focuses on an object can potentially reveal whether he/she has autism or visual impairment~\cite{tafasca2023childplay,chen2023early,mundy1990longitudinal,senju2009atypical}. In assisted driving scenarios, determining what object the driver is looking at can assess their attention and provide corresponding driving guidance~\cite{Chen_2023_ICCV, huang2022driver,lv2020improving}. In general, detecting the object gazed at by humans realizes the interaction between human sight and specific objects, fostering a richer semantic understanding of the scene.

\begin{figure}[!t]
\centering
\includegraphics[width=1\linewidth]{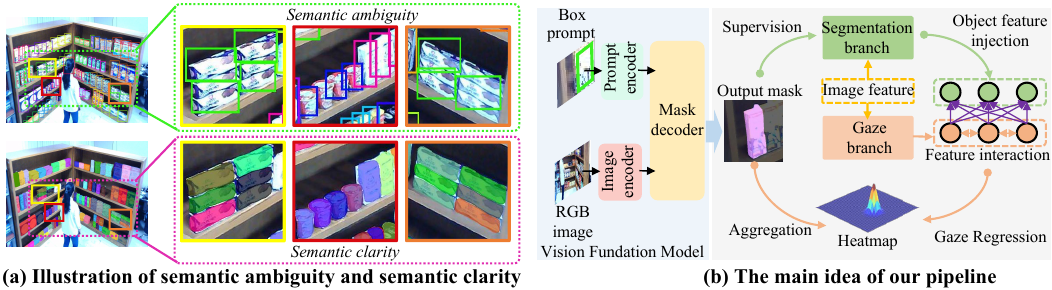}
\caption{
\textbf{(a)} Box-level supervision often fails to localize objects in dense settings precisely and leads to semantic ambiguity problems, whereas pixel-level supervision excels by providing clear semantic distinction through pixel-by-pixel predictions. \textbf{(b)} Vision foundation models can produce instance masks, thereby segmentation features can be used to improve the gaze regression branch's spatial perception, and the gaze object mask can help the heatmap focus on the gaze object.
}
\label{fig:intro}
\end{figure}

Tomas~\etal \cite{tomas2021goo} first proposed the gaze object detection task and the benchmark datasets GOO-Synth and GOO-Real, aiming to predict the object boxes and categories stared at by humans. Recently, Wang~\etal~\cite{wang2022gatector, wang24transgop} conducted an in-depth exploration of this task. Despite achieving promising results, they still suffer from the following limitations: \textbf{1)} They typically rely on box-level supervision, which is prone to suffer semantic ambiguity in dense object scenes. Especially when multiple objects are densely stacked or placed adjacent to each other, the box-level supervision algorithm may struggle to accurately separate them, as illustrated in Fig.~\ref{fig:intro}(a). 
\textbf{2)} The GOO~\cite{tomas2021goo} dataset only contains pixel-level annotations of synthetic images, 
leading to difficulties in generalizing to fine-grained predictions in real scenes.
\textbf{3)} They usually require additional head information as input, affecting the efficiency and flexibility of the model. While global modeling using the Transformer~\cite{tu2022end,tu2023gaze} can alleviate this problem, it lacks special attention to head features, which are crucial for gaze direction perception.
\textbf{4)} They directly fuse features to regress the gaze heatmap, lacking a full understanding of the spatial location and subtle detailed information of objects. 

Recently, vision foundational models (VFM) have gained significant attention due to their powerful generalization capabilities, which can achieve accurate pixel-level segmentation using a variety of prompts, \ie, boxes, points, or masks. 
In real scenes with small and densely packed objects, VFM can generate accurate boundaries and pixel-level masks (See Fig~\ref{fig:mask}) by utilizing boxes as prompt, which reduces semantic ambiguity and thus opens opportunities for gaze object segmentation. Motivated by this, we propose a more challenging task, gaze object segmentation (GOS). This task involves inferring the pixel-level mask of objects captured by human gaze behavior, helping to further enhance the understanding of human gaze behavior. Furthermore, we establish a GOS framework specifically for this task, which integrates the pixel-level supervision provided by VFM into gaze object prediction to alleviate the aforementioned limitations of previous methods~\cite{wang2022gatector, wang24transgop}. 

In the proposed model, we introduce the unified detection and segmentation Transformer, \ie MaskDINO~\cite{li2023mask}, to simultaneously obtain candidate gaze object boxes, segmentation masks, and head boxes of the subject, which are indispensable for enabling the model to infer in the real world and achieve accurate detection. Instead of requiring additional head input~\cite{wang2022gatector, wang24transgop} or ignoring head features by global modeling~\cite{tu2022end,tu2023gaze}, we propose a RoI reconstruction module to automatically obtain head region features based on holistic features and head box provided by the unified Transformer. This helps capture gaze direction information during gaze regression to improve model efficiency and flexibility. Subsequently, a space-to-object gaze regression strategy is developed to enhance the perception of the spatial location of objects and the understanding of detailed information. Specifically, inspired by the ability of humans to infer others' gaze objects, a dual attention fusion module is presented to establish an initial human-object gaze spatial connection. Then, we propose an object feature interaction module to refine the gaze spatial connection through semantically clear feature interaction between gaze regression features and object mask features, thus strengthening the modeling of the human-object gaze relationship. In the end, an energy aggregation loss is used to further guide the regression of the gaze heatmap. By focusing on pixel-level details and progressively improving the gaze representation, it further reduces semantic ambiguity, leading to the generation of more precise heatmaps for gazed objects.

Our main contributions could be summarized as follows:
\begin{itemize} 
    \item We introduce the vision foundation model into gaze object prediction to provide pixel-level supervision and propose a more challenging gaze object segmentation task, which focuses on identifying the pixel-level masks of objects captured by human gaze behavior.
    
    \item We present an all-in-one end-to-end framework for gaze object detection and segmentation without relying on additional head-related inputs, capable of simultaneously addressing gaze estimation, object detection, and segmentation. To the best of our knowledge, our work is the first to tackle gaze object prediction from the image segmentation perspective.
    
    \item We propose a space-to-object gaze regression approach that enhances the interaction between object and gaze branches, incrementally improving the modeling of the human-object gaze relationship and alleviating semantic ambiguity.
\end{itemize}

\section{Related Works}

\noindent \textbf{Vision Foundation Model.}
The vision foundation models (VFMs)~\cite{kirillov2023segment, zhao2023fast, ma2024segment} are typically built on large-scale datasets using self-supervised or semi-supervised methods, which have strong generalizability and zero-shot transfer capabilities. Segment Anything Model (SAM)~\cite{kirillov2023segment} received attention as a foundation model for image segmentation. Its portability enables it to be flexibly applied to various downstream tasks. Then, the medical foundation model MedSAM~\cite{ma2024segment} extends the foundation model to medical diagnosis. In this article, we choose SAM~\cite{kirillov2023segment} to obtain high-quality masks to boost gaze estimation and gaze object prediction.

\noindent \textbf{Gaze Object Detection.}
The gaze object detection is first proposed in ~\cite{tomas2021goo}, which aims to predict the box and categories of the object stared at by humans in an RGB image. Tomas~\etal \cite{tomas2021goo} released a benchmark dataset for GOP. Wang~\etal \cite{wang2022gatector} first proposed a GOP method named GaTector. However, GaTector requires additional head-related prior information, which hinders its application in the real world. Recently, the Transformer-based method~\cite{wang24transgop} achieves comparable performance. Although previous methods are effective to some extent, they can only provide instance-level box outputs, leading to potential ambiguities or misjudgments in scenarios with small and densely packed objects.

\noindent \textbf{Gaze Target Detection.} Early research on gaze focused on gaze direction estimation~\cite{zhang2015appearance, zhu2017monocular, wang2017real, zhang2017mpiigaze, cheng2018appearance, park2018deep, park2018learning, kellnhofer2019gaze360, wang2020learning, Chen_2023_ICCV, Cai_2023_CVPR, Ruzzi_2023_CVPR, Jin_2023_CVPR, cheng2022gaze}, which only predict the gaze vector and ignores the impact of the surrounding environment on the sight, resulting in limited application scenarios. Therefore, to capture higher-level gaze-related semantic information, researchers propose gaze target detection~\cite{recasens2017following, tonini2022multimodal,tonini2023object, lian2018believe,wang2023dual, chong2018connecting,chong2020detecting,bao2022escnet,gupta2022modular,miao2023patch,jin2022depth,hu2022gaze,li2021looking, hu2023gfie, tafasca2023childplay, 9740573}, which involves to infer the specific gaze point. Recasens~\etal \cite{recasens2015they} introduced the GazeFollow dataset and first proposed gaze target detection. Recently, Tu~\etal \cite{tu2022end, tu2023gaze} achieved end-to-end gaze target detection using Transformer, with a relatively low detection efficiency. Despite advancements in gaze target estimation, previous methods struggle with gaze regression due to spatial perception ambiguity. In contrast, our space-to-object regression strategy helps the model gradually converge to a certain gaze point, simplifying learning and enhancing detailed interaction with the object.

\begin{figure*}[!t]
\centering
\includegraphics[width=1\linewidth]{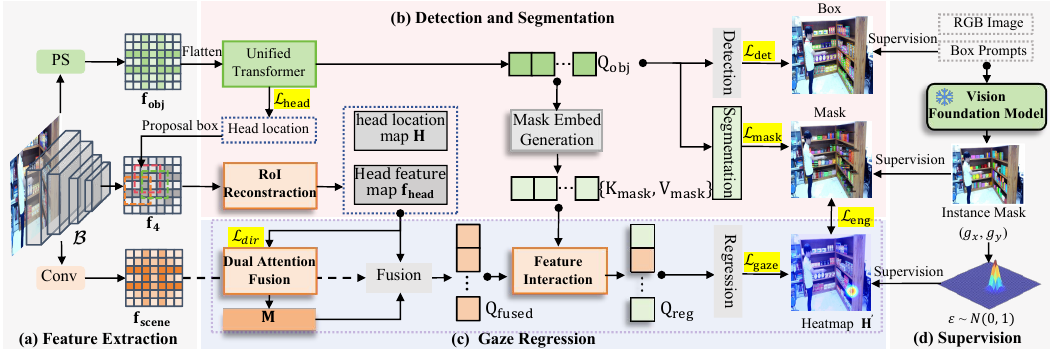}
\caption{\textbf{Overview of the proposed model.} \textbf{(a)} The feature extraction module extracts features for detection and regression branches. \textbf{(b)} The detection and segmentation branch identifies object and human head positions. \textbf{(c)} After obtaining head features, the gaze regression branch progressively refines its output: 1) employing a dual attention fusion module for initial human-object correlations; 2) leveraging a feature interaction module to incorporate semantically clear object-aware insights from the segmentation branch; 3) ultimately predicting the gaze heatmap. \textbf{(d)} Supervision signals are applied to both branches only during training.}
\label{fig:framework}
\end{figure*}

\section{Method} 
Given an RGB image, our model aims to precisely predict the pixel-level mask of the object humans are looking at. We first present the overall structure of our method, then introduce the VFM-based mask supervision generation method, and finally describe each module of the proposed model in detail.
\subsection{Overview}
\label{overview}
As shown in Fig.~\ref{fig:framework}, our method consists of four components: a feature extractor for specific feature extraction, a detection and segmentation module predicting candidate gaze objects boxes, masks, and human head location, a gaze regressor producing gaze object heatmap with a space-to-object strategy, and a supervision information generation module creating pixel-level masks and ground truth heatmap during training. At training, the overall loss function is defined as:
\begin{equation}
  \label{eq:loss}
  \mathcal{L} = \mathcal{L}_{\rm det} + \alpha\mathcal{L}_{\rm  dir} +\beta\mathcal{L}_{\rm gaze} + \gamma\mathcal{L}_{\rm eng}
\end{equation}
where $\alpha$, $\beta$, and $\gamma$ are the weights of the gaze direction loss, gaze heatmap loss, and energy aggregation loss respectively. Before model training, we choose SAM to generate pixel-level masks for the objects. See Sec.~\ref{sec:mask_generation} for details. 

\noindent \textbf{Feature extraction.}
As shown in \ref{fig:framework}(a), given an RGB image $\mathcal{I}\in \mathbb{R}^{3\times H\times W}$, we first use the backbone $\mathcal{B}$ to extract multi-scale features $\{{\mathbf{f}_{ 1}, \mathbf{f}_{ 2}, \mathbf{f}_{ 3}, \mathbf{f}_{ 4}}\}$. In previous work~\cite{wang2022gatector}, pixel shuffle~\cite{shi2016real} significantly improved small object detection by enlarging feature maps for detailed feature capture. Therefore, we integrate it into dense prediction tasks, \ie, image segmentation. In our model, the multi-scale features $\{{\mathbf{f}_{ 1}, \mathbf{f}_{ 2}, \mathbf{f}_{ 3}, \mathbf{f}_{ 4}}\}$ are fed into the pixel shuffle module to generate object-specific features $\mathbf{f}_{ i}^{ obj} =  \mathcal{\phi}^{ obj} (\mathbf{f}_{ i}, \eta)$, where $\mathbf{f}_{i}$ denotes the $i$-th feature layer, $\eta$ denotes the scale factor, $\mathcal{\phi}^{obj}(\cdot)$ denotes the pixel shuffle module, which consists of a pixel shuffle operation and a convolution layer that transforms the feature channels to be consistent with the original features of the unified Transformer~\cite{li2023mask}. Finally, the scene saliency features $\mathbf{f}_{\rm scene} \in \mathbb{R}^{1024 \times 7 \times 7}$ are extracted by a scene residual block using high dimensional features $\mathbf{f}_{\rm 4} \in \mathbb{R}^{2048 \times 7 \times 7}$.

\noindent \textbf{Unified detection and segmentation Transformer.}
In this paper, we employ a unified Transformer for candidate gaze object detection, segmentation, and human head positioning, integrating MaskDINO~\cite{li2023mask} for candidate gaze object detection,  segmentation, and the generation of pixel-level mask feature embeddings and a head decoder for head box regression, which facilitate accurate head-related feature extraction in head feature reconstruction module and accurate prediction in real-world settings. The training loss is defined as:
\begin{equation}
  \label{eq:loss_det}
  \mathcal{L}_{\rm det} =  \mathcal{L}_{\rm det}^{\rm obj} + \mathcal{L}_{\rm mask}^{\rm obj} + \mathcal{L}_{\rm  det}^{\rm head}
\end{equation}
$\mathcal{L}_{\rm det}^{\rm obj}$ and $\mathcal{L}_{\rm mask}^{\rm obj}$ is the object detection and segmentation loss respectively. $\mathcal{L}_{\rm det}^{\rm head}$ is the head decoder loss, which is consistent with the $\mathcal{L}_{\rm det}^{\rm obj}$. See Sec.~\ref{sec:detection_branch} for details.

\noindent \textbf{Space-to-object gaze regression.}
We employ a space-to-object regression strategy for gaze heatmap prediction. The module begins by representing human gaze behavior with a gaze vector. Subsequently, a dual attention fusion module establishes an initial human-object spatial connection. Finally, pixel-level object location knowledge from the mask branch is introduced to achieve semantically clear feature interaction, refining the gaze heatmap regression while alleviating semantic ambiguity. Optimization involves $\mathcal{L}_{\rm dir}$ and $\mathcal{L}_{\rm gaze}$ for gaze direction and gaze heatmap, respectively. The energy aggregation loss $\mathcal{L}_{\rm eng}$ guides the gaze heatmap to focus on the gaze object mask. See Sec.~\ref{sec:gaze_branch} for details.

\begin{figure}[!t]
\centering
\includegraphics[width=1\linewidth]{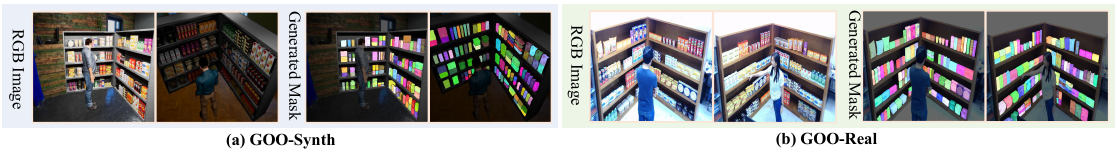}
\caption{Illustration of instance mask generated by VFM.}
\label{fig:mask}
\end{figure}

\subsection{Instance mask generation via VFM.}
\label{sec:mask_generation}
Given an RGB image $\mathcal{I}$, we choose the vision foundation model SAM~\cite{kirillov2023segment} to obtain instance masks for model training. Specifically, image $\mathcal{I}$ and prompt $\mathcal{P}_{box}$ are first fed into the image encoder $\Psi_{Image}$ and prompt encoder $\Psi_{prompt}$, respectively. Then, the mask $\mathcal{P}_{mask}$ and confidence $\mathcal{S}$ are output by the mask decoder $\Phi_{mask}$, which can be formulated as: 
\begin{equation}
  { \mathcal{P}_{\rm mask}, \mathcal{S}} = \Phi_{mask} (\Psi_{Image}(\mathcal{I}), \Psi_{prompt}(\mathcal{P}_{box}))
\end{equation}
Instance segmentation by using boxes produces accurate masks for most objects, but accurate segmentation of objects with unclear texture or edge features remains challenging. Therefore, we segment again using $\mathcal{P}_{mask}$ as a dense prompt to produce final mask $\mathcal{M}_{\rm mask}$, which can be formulated as: 
\begin{equation}
  { \mathcal{M}_{\rm mask}, \mathcal{S}} = \Phi_{mask} (\Psi_{Image}(\mathcal{I}), \Psi_{prompt}(\mathcal{P}_{mask}))
\end{equation}
Finally, $\mathcal{M}_{mask}$ is used to train the gaze object segmentation model. The generated masks are shown in Fig.~\ref{fig:mask}.

\subsection{Unified detection and segmentation Transformer}
\label{sec:detection_branch}
The unified detection and segmentation Transformer aims to detect candidate gaze objects and locate human heads while extracting high-level pixel mask features, which comprises a detection and segmentation Transformer, a head decoder, and a RoI feature reconstruction module. As shown in Fig.~\ref{fig:framework}(b), the object-specific features $\{{\mathbf{f}_{ 1}^{ obj}, \mathbf{f}_{ 2}^{ obj}, \mathbf{f}_{ 3}^{ obj}, \mathbf{f}_{ 4}^{ obj}}\}$ are fed it to detect object boxes, categories, masks, and human head boxes.

\noindent \textbf{Object detection and segmentation}
The global attention of Transformer ensures powerful long-distance modeling and global perception, achieving excellent detection performance through layer-by-layer self-attention refinement. We use MaskDINO~\cite{li2023mask} as a foundational detection and segmentation structure due to its robust representation capabilities and fast convergence speed.

\noindent \textbf{Head decoder}
Our model is designed for real-world gaze object detection and segmentation without relying on head priors. We introduce a head decoder based on the unified Transformer, using a 3-layer self-attention decoder to generate proposals for head-related feature reconstruction. During training, we filter head queries with IoU > 0.5 as positives. During inference, boxes are selected based on confidence. Similar to other DETR-like detectors~\cite{zhang2022dino, zhu2021deformable}, the head decoder uses cross-entropy, $L_{1}$, and GIoU loss for classification and box regression.

\noindent \textbf{Head feature reconstruction}
Previous methods typically extract head features from manually cropped head images, limiting model inference efficiency and flexibility in the real world. To this end, we design a RoI feature reconstruction module to reconstruct head-related features based on holistic scene features and the head box from the unified Transformer. In this module, we first use the head bounding box ${R = (x_1, y_1, x_2, y_2)}$, which represents the upper left and lower right corner coordinates, as the region proposal to crop the regions of interest (ROIs). Then, those candidate ROIs are refined by ROIAlign~\cite{he2017mask}, which maps the head region to the corresponding pixels in the holistic image features $\mathbf{f}_{\rm 4} \in \mathbb{R}^{2048 \times 7 \times 7}$ and produces a fixed-size feature map $\mathbf{f}_{\rm head} \in \mathbb{R}^{2048 \times 7 \times 7}$. For head location map generation, following previous works~\cite{recasens2015they, chong2020detecting, gupta2022modular, wang2022gatector} to generate a binary image, setting the head area to 1 and the rest to 0. Additionally, a specific residual block extracts gaze saliency features $\mathbf{f}_{\rm gaze} \in \mathbb{R}^{1024 \times 7 \times 7}$, further enhancing the model's attention to the gaze direction.

\subsection{Space-to-object gaze regression}
\label{sec:gaze_branch}

We propose a space-to-object gaze regression approach to gradually refine the human-object gaze relationship. Fig.~\ref{fig:framework}(c) shows our gaze regression branch with a dual attention fusion module establishing an initial human-object spatial connection and a feature interaction module strengthening this connection using semantically clear feature interaction.

\begin{figure}[!t]
\centering
\includegraphics[width=0.9\linewidth]{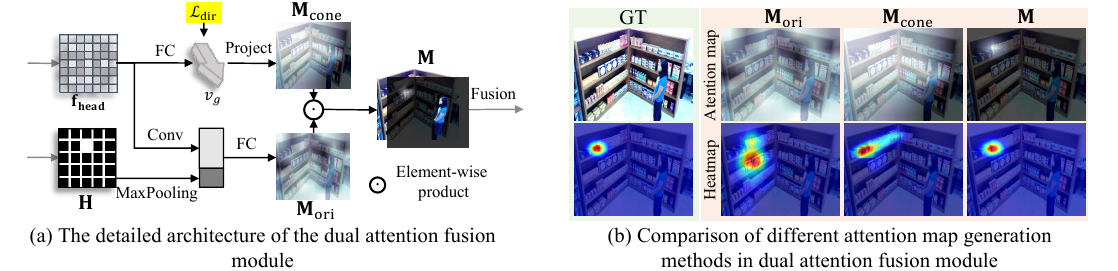}
\caption{Illustration of spatial perception for gaze objects.}
\label{fig:chor_module}
\end{figure}

\noindent \textbf{Spatial perception for gaze objects.}
When humans infer what another person is staring at, they usually first observe the gaze direction of the others and search for the gazed object along this direction. Inspired by this, we propose the dual attention fusion module for the spatial perception of gaze objects, establishing an initial human-object spatial connection.

As shown in Fig.~\ref{fig:chor_module}(a), we first use global average pooling to obtain the global features of $\mathbf{f}_{\rm head}$, followed by fully connected layers to transform the features into a 2D gaze vector $\mathbf{v}_{\rm g}$. To achieve spatial perception, we project it into a gaze cone map $\mathbf{M}_{\rm cone} \in \mathbb{R}^{1 \times 7 \times 7}$, activating the area around the gaze direction:
\begin{equation}
\mathbf{M}^{\rm  cone}_{\rm i,j} = Max(0, \cos (\mathbf{v}_{\rm g}, \mathbf{v}_{\rm i,j}))
\end{equation}
where $\mathbf{v}_{\rm g}$ and $\mathbf{v}_{\rm i,j}$ respectively denotes the predicted gaze vector and the vector composed of the eye point and each point on the gaze field.
During training, $\mathbf{v}_{\rm g}$ is optimized by $\mathcal{L}_{\rm dir}$ to obtain a more accurate gaze direction and activation map. $\mathcal{L}_{\rm dir}$ is defined as follows:
\begin{equation}
\mathcal{L}_{\rm  dir} = 1 - \sum_{i=1}^{n} (\mathbf{v}_{\rm g}^{i} \cdot \mathbf{v}_{\rm t}^{i})
\end{equation}
where vector $\mathbf{v}_{\rm g}$ and $\mathbf{v}_{\rm t}$ denote the predicted gaze direction and ground truth, respectively. 
During backpropagation, $\mathcal{L}_{\rm dir}$ adjusts parameters of head feature $\mathbf{f}
_{\rm head}$, providing more direction information for gaze regression. To boost scene attention, we adopt the method from GaTector~\cite{wang2022gatector} to generate an original attention map $\mathbf{M}_{\rm ori} \in \mathbb{R}^{1 \times 7 \times 7}$, activating scene salience area based on head properties. By multiplying $\mathbf{M}_{\rm cone}$ with $\mathbf{M}_{\rm ori}$ via element-wise dot product, we generate the final spatial perception map $\mathbf{M}$:
\begin{equation}
\mathbf{M} = \mathbf{M}_{\rm  ori} \odot \mathbf{M}_{\rm  cone}
\end{equation}
This compels the model to search spaces more likely gazed at by humans along gaze direction, establishing an initial human-object gaze spatial relationship.

\noindent\textbf{Semantically clear feature interaction.}
Previous methods~\cite{wang2022gatector, wang24transgop} usually directly fuse features to regress the gaze heatmap based on spatial saliency, lacking interaction with object details and facing challenges in accurately predicting heatmap for the gazed object. In our method, we introduce a feature interaction module, which leverages semantically clear mask knowledge from VFM to achieve feature interaction, refining human-object gaze relationship modeling. 

As shown in Fig.~\ref{fig:framework}(c), we initially extract detailed mask features from the mask branch. Subsequently, the fused feature $\mathbf{f}_{\rm fuse}$ is utilized for pixel-level attention perception. Given that candidate gaze objects can be located anywhere in the image, the Transformer's global modeling excels in accomplishing this task. Therefore, utilize a transformer layer, akin to DETR~\cite{carion2020end}, to capture the global relationship between human gaze behavior and object masks. For fine-grained object mask features, a three-layer MLP is applied to process the decoder query $\mathcal{Q}_{\rm obj}$ of the unified Transformer, generating mask embeddings:
\begin{equation}
  \mathcal{Q}_{\rm mask} = \mathbf{MLP} (\mathcal{Q}_{\rm obj})
\end{equation}
where the $\mathcal{Q}_{\rm mask}$ denote the generated mask embedding. We first apply a layer self-attention for $\mathbf{f}_{\rm fuse}$:
\begin{equation}
  \mathbf{f}_{\rm e} = \mathbf{FFN}(\mathbf{SelfAttn} (\mathbf{f}_{\rm fuse}))
\end{equation}
where $\mathbf{f}_{\rm fuse}$, $ \mathcal{\rm SelfAttn}(\cdot)$, and $\mathbf{f}_{\rm e}$ denote the fused feature $\mathbf{f}_{\rm fuse}$, self-attention operation, and encoded $\mathbf{f}_{\rm fuse}$, respectively. Then, the $\mathcal{Q}_{\rm mask}$ is perceived by the $\mathbf{f}_{\rm e}$ as key-value pairs, \ie, $\{\mathbf{K}_{\rm mask}, \mathbf{V}_{\rm mask}\}$:
\begin{equation}
  \mathbf{f}_{\rm reg} = \mathbf{FFN}(\mathbf{CrossAttn} (\mathbf{f}_{\rm e}, \{\mathbf{K}_{\rm mask}, \mathbf{V}_{\rm mask}\}))
\end{equation}
where $\mathbf{f}_{\rm reg}$ and $\mathbf{\rm CrossAttn}(q, kv)$ denote the regression feature and cross-attention operation, respectively. This operation refines the perception of the object by providing a pixel-level understanding of the gaze object in each token of the regression feature. It effectively alleviates semantic ambiguity and strengthens the ability of the model to represent the human-object gaze relationship.

\noindent\textbf{Ultimate gaze regression.} Finally, following previous work~\cite{wang2022gatector} to generate the ground truth gaze heatmap and use $L_{2}$ loss to supervise gaze heatmap regression. $\mathcal{L}_{\rm eng}$ proposed by~\cite{wang24transgop} guides gaze heatmap energy aggregation towards the gaze object. We use the gaze object mask for more accurate error measurements. Space-to-object gaze regression approach gradually refines the human-object gaze behavior representation, regressing a more accurate gaze heatmap.

\section{Experiments}
\subsection{Experiments Settings}
Datasets and implementation details are given in \textbf{supplementary material}.

\noindent \textbf{Metrics:} 
Following previous work~\cite{wang2022gatector, wang24transgop}, we use AUC (Area Under the Curve), L2 distance, and Angle error for gaze estimation evaluation, and mSoC (min Single over Closure)\footnote{\noindent mSoC metric proposed by Wang~\etal. URL: \url{https://arxiv.org/abs/2112.03549}.} under different thresholds for gaze object prediction evaluation. Following MaskDINO~\cite{li2023mask}, AP (Average Precision) is used for instance segmentation and object detection evaluation. To evaluate the gaze object segmentation, we use masks to replace boxes when calculating mSoC (details refer to the \textbf{supplementary material}). For a fair comparison, we reported results for both real-world and non-real-world settings. The non-real-world settings involve inputting additional head images and head location maps while real-world settings only use scene images. Furthermore, to compare with GTR~\cite{tu2023gaze}, we also adopt the evaluation method proposed by ~\cite{tu2023gaze}~\footnote{\noindent Evaluation method proposed by Tu~\etal, URL: \url{https://ieeexplore.ieee.org/document/10262016}.}. This method computes gaze estimation metrics only for results with L2 distance < 0.15 and head box IoU > 0.5, which is an incomplete evaluation for AUC, L2 distance, and Angle error.

\begin{table}[!t]
  \centering
  \caption{Gaze object detection results on GOO-Synth and GOO-Real datasets. $\ast$ indicates using additional head detector to achieve real-world settings.}
  \resizebox{0.95\linewidth}{!}{
    \begin{tabular}{c|l|cccc|cccc|cccc|c|c}
    \toprule
    \multicolumn{2}{c|}{\multirow{2}[2]{*}{Method}} & \multicolumn{4}{c|}{GOO-Synth} & \multicolumn{4}{c|}{GOO-Real No Pretrain} & \multicolumn{4}{c|}{GOO-Real Pretrain} & \multirow{2}[2]{*}{Params} & \multirow{2}[2]{*}{FPS} \\
    \cmidrule{3-14}
    \multicolumn{2}{c|}{} & mSoC  & 50 & 75 & 95 & mSoC  & 50 & 75 & 95 & mSoC  & 50 & 75 & 95 &       &  \\
    \midrule
    \multirow{3}[4]{*}{\rotatebox{90}{Non-Real}\quad} & GaTector~\cite{wang2022gatector} & 67.94  & 98.14  & 86.25  & 0.12  & 62.40  & 95.10  & 73.50  & 0.20  & 71.20  & 97.50  & 88.80  & 2.70  & 60.78M & 14.1  \\
          & TransGOP~\cite{wang24transgop} & 92.80  & 99.00  & 98.50  & 51.90  & 82.60  & 97.80  & 89.40  & 6.50  & 89.00  & 98.90  & 97.50  & 33.20  & 94.03M & 13.7  \\
\cmidrule{2-16}          & \cellcolor[rgb]{ .906,  .902,  .902}Ours & \cellcolor[rgb]{ .906,  .902,  .902}91.19  & \cellcolor[rgb]{ .906,  .902,  .902}93.97  & \cellcolor[rgb]{ .906,  .902,  .902}93.46  & \cellcolor[rgb]{ .906,  .902,  .902}73.27  & \cellcolor[rgb]{ .906,  .902,  .902}81.75  & \cellcolor[rgb]{ .906,  .902,  .902}98.78  & \cellcolor[rgb]{ .906,  .902,  .902}96.43  & \cellcolor[rgb]{ .906,  .902,  .902}7.13  & \cellcolor[rgb]{ .906,  .902,  .902}81.89  & \cellcolor[rgb]{ .906,  .902,  .902}98.99  & \cellcolor[rgb]{ .906,  .902,  .902}96.63  & \cellcolor[rgb]{ .906,  .902,  .902}7.28  & \cellcolor[rgb]{ .906,  .902,  .902}80.99M & \cellcolor[rgb]{ .906,  .902,  .902}14.5  \\
    \midrule
    \multirow{3}[4]{*}{\rotatebox{90}{Real}\quad} & GaTector~\cite{wang2022gatector}$\ast$ & 56.19  & 88.08  & 66.18  & -     & 40.87  & 82.05  & 33.54  & 0.30  & 66.40  & 96.22  & 80.49  & 0.40  & 75.73M & 9.3  \\
          & TransGOP~\cite{wang24transgop}$\ast$ & 91.60  & 99.00  & 98.10  & 48.70  & 79.70  & 97.80  & 91.40  & 11.20  & 83.20  & 98.80  & 95.30  & 14.40  & 106.37M & 11.6  \\
\cmidrule{2-16}          & \cellcolor[rgb]{ .906,  .902,  .902}Ours & \cellcolor[rgb]{ .906,  .902,  .902}90.77  & \cellcolor[rgb]{ .906,  .902,  .902}93.92  & \cellcolor[rgb]{ .906,  .902,  .902}93.45  & \cellcolor[rgb]{ .906,  .902,  .902}69.58  & \cellcolor[rgb]{ .906,  .902,  .902}81.35  & \cellcolor[rgb]{ .906,  .902,  .902}98.95  & \cellcolor[rgb]{ .906,  .902,  .902}96.02  & \cellcolor[rgb]{ .906,  .902,  .902}2.61  & \cellcolor[rgb]{ .906,  .902,  .902}81.60  & \cellcolor[rgb]{ .906,  .902,  .902}99.03  & \cellcolor[rgb]{ .906,  .902,  .902}96.54  & \cellcolor[rgb]{ .906,  .902,  .902}6.03  & \cellcolor[rgb]{ .906,  .902,  .902}84.26M & \cellcolor[rgb]{ .906,  .902,  .902}14.1  \\
    \bottomrule
    \end{tabular}%
    }
  \label{tab:god}%
\end{table}%

\begin{table*}[!t]
  \centering
  \caption{Ours gaze object segmentation results on GOO-Real and GOO-Synth.}
  \resizebox{0.95 \linewidth}{!}{
    \begin{tabular}{l|cccc|cccc|cccc|c|c}
    \toprule
    \multicolumn{1}{c|}{\multirow{2}[4]{*}{Settings}} & \multicolumn{4}{c|}{GOO-Synth} & \multicolumn{4}{c|}{GOO-Real No Pretrain} & \multicolumn{4}{c|}{GOO-Real Pretrain} & \multirow{2}[4]{*}{Params} & \multirow{2}[4]{*}{FPS} \\
\cmidrule{2-13}          & mSoC  & 50 & 75 & 95 & mSoC  & 50 & 75 & 95 & mSoC  & 50 & 75 & 95 &       &  \\
    \midrule
    Non-Real & \cellcolor[rgb]{ .906,  .902,  .902}82.68  & \cellcolor[rgb]{ .906,  .902,  .902}93.70  & \cellcolor[rgb]{ .906,  .902,  .902}91.68  & \cellcolor[rgb]{ .906,  .902,  .902}26.83  & \cellcolor[rgb]{ .906,  .902,  .902}86.39  & \cellcolor[rgb]{ .906,  .902,  .902}98.95  & \cellcolor[rgb]{ .906,  .902,  .902}97.44  & \cellcolor[rgb]{ .906,  .902,  .902}20.07  & \cellcolor[rgb]{ .906,  .902,  .902}89.39  & \cellcolor[rgb]{ .906,  .902,  .902}98.80  & \cellcolor[rgb]{ .906,  .902,  .902}98.02  & \cellcolor[rgb]{ .906,  .902,  .902}33.03  & \cellcolor[rgb]{ .906,  .902,  .902}80.99M & \cellcolor[rgb]{ .906,  .902,  .902}14.5 \\
    Real & \cellcolor[rgb]{ .906,  .902,  .902}82.73  & \cellcolor[rgb]{ .906,  .902,  .902}93.67  & \cellcolor[rgb]{ .906,  .902,  .902}91.79  & \cellcolor[rgb]{ .906,  .902,  .902}26.85  & \cellcolor[rgb]{ .906,  .902,  .902}85.94  & \cellcolor[rgb]{ .906,  .902,  .902}98.99  & \cellcolor[rgb]{ .906,  .902,  .902}97.29  & \cellcolor[rgb]{ .906,  .902,  .902}16.70  & \cellcolor[rgb]{ .906,  .902,  .902}86.96  & \cellcolor[rgb]{ .906,  .902,  .902}98.94  & \cellcolor[rgb]{ .906,  .902,  .902}97.75  & \cellcolor[rgb]{ .906,  .902,  .902}22.21  & \cellcolor[rgb]{ .906,  .902,  .902}84.26M & \cellcolor[rgb]{ .906,  .902,  .902}14.1 \\
    \bottomrule
    \end{tabular}%
    }
  \label{tab:gos}%
  
\end{table*}%

\begin{figure}[!t]
\centering
\includegraphics[width=0.9\linewidth]{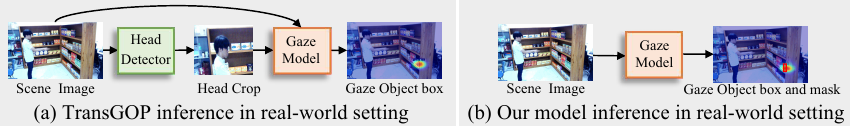}
\caption{Comparison of TransGOP~\cite{wang24transgop} and our method inference in the real world.}
\label{fig:experiments}
\end{figure}

\subsection{Comparison with SOTA}

\noindent \textbf{Gaze object detection and segmentation.}
Table~\ref{tab:god} shows the performance of gaze object detection on GOO-Synth and GOO-Real. In real-world settings, previous methods~\cite{wang2022gatector, wang24transgop} significantly increase parameters and slow down the inference speed due to the additional head detector (As shown in Fig.~\ref{fig:experiments} (a)) and they can only achieve box-level prediction. In contrast, although our method is slightly inferior to TransGOP~\cite{wang24transgop} in detection performance, our model runs significantly more efficiently because it integrates the acquisition of head features into the overall framework (As shown in Fig.~\ref{fig:experiments} (b)) and can output bounding boxes and segmentation masks simultaneously. The above results also show that our space-to-object gaze regression method can effectively model the human-object gaze relationship. Table~\ref{tab:gos} shows our gaze object segmentation performance in real-world and non-real-world scenarios. On GOO-Synth, our model achieves 82.68\% and 82.73\% mSoC in real-world and non-real-world settings. On GOO-Real, without pre-training, our model performs well in both scenarios (86.39\% mSoC and 85.94\% mSoC). After pre-training, it reaches 89.39\% mSoC and 86.96\% mSoC. The above experimental results show that the pixel-level mask knowledge provided by VFMs can effectively alleviate semantic ambiguity while achieving pixel-level prediction. Even under the multi-task learning framework, we still achieve an FPS of 14.1. This is mainly attributed to the automatic incorporation of head features, avoiding the need for an additional head detector and thus enhancing the efficiency of the model.

\noindent \textbf{Object detection and segmentation.}
Tables~\ref{tab:segmentation} and~\ref{tab:detection} show our model's instance segmentation and object detection performance. For a fair comparison, we adjusted the input size of MaskDINO~\cite{li2023mask} to match ours (224 × 224). In Table~\ref{tab:segmentation}, our model outperforms MaskDINO by 8.66 AP on GOO-Real (70.85 vs. 79.51) and by 5.81 AP on GOO-Synth (77.05 vs. 82.86) in real-world scenarios, with similar advantages in non-real-world settings of our model. These results show the challenge faced by the original MaskDINO~\cite{li2023mask} in capturing detailed features in scenes with dense objects and the effectiveness of the pixel shuffle of our model in capturing detailed features. Table~\ref{tab:detection} further demonstrates the superiority of our model in object detection over previous methods and MaskDINO~\cite{li2023mask}.

\begin{table}[!t]
  \centering
  \caption{Instance segmentation results on GOO-Synth and GOO-Real datasets}
  \resizebox{0.85\linewidth}{!}{
    \begin{tabular}{c|l|cccc|cccc|c}
    \toprule
    \multicolumn{2}{c|}{\multirow{2}[4]{*}{Method}} & \multicolumn{4}{c|}{GOO-Synth Dataset} & \multicolumn{4}{c|}{GOO-Real Dataset} & \multirow{2}[4]{*}{Params} \\
\cmidrule{3-10}    \multicolumn{2}{c|}{} &\quad AP \quad   &\quad 50 \quad&\quad 75 \quad&\quad  95 \quad&\quad AP \quad&\quad 50 \quad&\quad 75 \quad&\quad 95  \\
    \midrule
    \multicolumn{1}{l|}{Non-Real} & \cellcolor[rgb]{ .906,  .902,  .902}Ours  & \cellcolor[rgb]{ .906,  .902,  .902}79.33  & \cellcolor[rgb]{ .906,  .902,  .902}93.51  & \cellcolor[rgb]{ .906,  .902,  .902}91.09  & \cellcolor[rgb]{ .906,  .902,  .902}11.72  & \cellcolor[rgb]{ .906,  .902,  .902}82.91  & \cellcolor[rgb]{ .906,  .902,  .902}98.85  & \cellcolor[rgb]{ .906,  .902,  .902}95.09  & \cellcolor[rgb]{ .906,  .902,  .902}11.22  & \cellcolor[rgb]{ .906,  .902,  .902}80.99M \\
    \midrule
    \multirow{2}[4]{*}{Real} & MaskDINO~\cite{li2023mask} & 70.85  & 95.54  & 83.31  & 1.81  & 77.05  & 98.38  & 90.45  & 4.26  & 52.03M \\
\cmidrule{2-11}          & \cellcolor[rgb]{ .906,  .902,  .902}Ours  & \cellcolor[rgb]{ .906,  .902,  .902}78.56  & \cellcolor[rgb]{ .906,  .902,  .902}96.00  & \cellcolor[rgb]{ .906,  .902,  .902}90.56  & \cellcolor[rgb]{ .906,  .902,  .902}8.43  & \cellcolor[rgb]{ .906,  .902,  .902}82.86  & \cellcolor[rgb]{ .906,  .902,  .902}99.02  & \cellcolor[rgb]{ .906,  .902,  .902}95.57  & \cellcolor[rgb]{ .906,  .902,  .902}10.09  & \cellcolor[rgb]{ .906,  .902,  .902}84.26M \\
    \bottomrule
    \end{tabular}%
    }
  \label{tab:segmentation}%
\end{table}%

\begin{table}[!t]
  \centering
  \caption{Object detection results on GOO-Synth and GOO-Real datasets. $\ast$ indicates using additional head detector to achieve real-world settings.}
  \resizebox{0.85\linewidth}{!}{
    \begin{tabular}{c|l|cccc|cccc|c}
    \toprule
    \multicolumn{2}{c|}{\multirow{2}[4]{*}{Method}} & \multicolumn{4}{c|}{GOO-Synth Dataset} & \multicolumn{4}{c|}{GOO-Real Dataset} & \multirow{2}[4]{*}{Params} \\
\cmidrule{3-10}    \multicolumn{2}{c|}{}  &\quad AP    \quad&\quad 50 \quad&\quad 75 \quad&\quad  95 \quad&\quad AP \quad&\quad 50 \quad&\quad 75 \quad&\quad 95 &  \\
    \midrule
    \multirow{3}[4]{*}{Non-Real} & GaTector~\cite{wang2022gatector} & 56.80  & 95.30  & 62.50  & 0.10  & 52.25  & 91.92  & 55.34  & 0.10  & 60.78M \\
          & TransGOP~\cite{wang24transgop} & 87.60  & 99.00  & 97.30  & 25.50  & 77.20  & 97.80  & 89.40  & 6.50  & 94.03M \\
\cmidrule{2-11}          & \cellcolor[rgb]{ .906,  .902,  .902}Ours & \cellcolor[rgb]{ .906,  .902,  .902}88.33  & \cellcolor[rgb]{ .906,  .902,  .902}93.88  & \cellcolor[rgb]{ .906,  .902,  .902}93.08  & \cellcolor[rgb]{ .906,  .902,  .902}50.70  & \cellcolor[rgb]{ .906,  .902,  .902}77.32  & \cellcolor[rgb]{ .906,  .902,  .902}98.65  & \cellcolor[rgb]{ .906,  .902,  .902}93.91  & \cellcolor[rgb]{ .906,  .902,  .902}1.93  & \cellcolor[rgb]{ .906,  .902,  .902}80.99M \\
    \midrule
    \multirow{4}[4]{*}{Real} & GaTector~\cite{wang2022gatector}$\ast$ & 43.50  & 86.23  & 37.52  & -     & 29.21  & 74.02  & 16.13  & -     & 75.73M \\
          & TransGOP~\cite{wang24transgop}$\ast$ & 86.00  & 98.80  & 96.20  & 23.80  & 73.30  & 96.70  & 86.00  & 4.10  & 106.37M \\
          & MaskDINO~\cite{li2023mask} & 72.93  & 96.26  & 88.13  & 2.04  & 74.44  & 98.72  & 91.19  & 0.19  & 52.03M \\
          \cmidrule{2-11} & \cellcolor[rgb]{ .906,  .902,  .902}Ours & \cellcolor[rgb]{ .906,  .902,  .902}87.60  & \cellcolor[rgb]{ .906,  .902,  .902}93.72  & \cellcolor[rgb]{ .906,  .902,  .902}73.07  & \cellcolor[rgb]{ .906,  .902,  .902}45.53  & \cellcolor[rgb]{ .906,  .902,  .902}75.20  & \cellcolor[rgb]{ .906,  .902,  .902}98.94  & \cellcolor[rgb]{ .906,  .902,  .902}93.20  & \cellcolor[rgb]{ .906,  .902,  .902}0.76  & \cellcolor[rgb]{ .906,  .902,  .902}84.26M \\
    \bottomrule
    \end{tabular}%
    }
  \label{tab:detection}%
\end{table}%

\noindent \textbf{Gaze estimation.}
Tables~\ref{tab:ge_synth} and~\ref{tab:ge_real} show the gaze estimation performance on GOO-Synth and GOO-Real, respectively. In Table~\ref{tab:ge_synth}, our method outperforms GTR~\cite{tu2023gaze} and TransGOP~\cite{wang24transgop} by 23.9 mAP and 10.6 mAP, respectively. Even directly evaluating all prediction results in the real world, our model reduces L2 distance and angle errors compared to TransGOP~\cite{wang24transgop} by 0.015 and 1.8, respectively. Applying non-real-world methods to the real world significantly degrades performance, highlighting our real-world model is more efficient. Table~\ref{tab:ge_real} verifies our method on GOO-Real, achieving state-of-the-art results without pre-training. Even with direct inference in a real-world setting, our method outperforms previous non-real-world methods. After pre-training, our model exhibits more pronounced performance advantages. These results show the effectiveness of our space-to-object gaze regression method in modeling human-object gaze relationships while achieving more accurate gaze heatmap regression.

\subsection{Ablation study and model analysis} 

\begin{figure}[!t]
\begin{minipage}[!b]{0.47\textwidth}
\makeatletter\def\@captype{table}
\centering
\caption{Gaze estimation results on GOO-Synth dataset. $\dagger$ indicates the evaluation method proposed by~\cite{tu2023gaze}.}
\resizebox{1\linewidth}{!}{
    \begin{tabular}{c|l|cccc}
    \toprule
    \multicolumn{2}{c|}{Method} & AUC    & Dist.$\downarrow$ & Ang.$\downarrow$  & mAP  \\
    \midrule
    \multirow{7}[4]{*}{\rotatebox{90}{Non-Real}\quad} & GazeFollow~\cite{recasens2015they} & 0.929  & 0.162  & 33.0  & - \\
          & Lian~\cite{lian2018believe}  & 0.954  & 0.107  & 19.7  & - \\
          & VideoAttention~\cite{chong2020detecting} & 0.952  & 0.075  & 15.1  & - \\
          & GaTector~\cite{wang2022gatector} & 0.957  & 0.073  & 14.9  & - \\
          & GTR~\cite{tu2023gaze}   & 0.960  & 0.071  & 14.5  & - \\
          & TransGOP~\cite{wang24transgop} & 0.963  & 0.079  & 13.3  & - \\
\cmidrule{2-6}          & \cellcolor[rgb]{ .906,  .902,  .902}Ours & \cellcolor[rgb]{ .906,  .902,  .902}0.947  & \cellcolor[rgb]{ .906,  .902,  .902}0.072  & \cellcolor[rgb]{ .906,  .902,  .902}14.8  & \cellcolor[rgb]{ .906,  .902,  .902}- \\
    \midrule
    \multirow{8}[4]{*}{\rotatebox{90}{Real}\quad} & GazeFollow~\cite{recasens2015they}$\dagger$ & 0.832  & 0.317  & 42.6  & 0.468  \\
          & Lian~\cite{lian2018believe}$\dagger$  & 0.903  & 0.153  & 28.1  & 0.454  \\
          & VideoAttention~\cite{chong2020detecting}$\dagger$ & 0.912  & 0.143  & 24.5  & 0.489  \\
          & GaTector~\cite{wang2022gatector}$\dagger$ & 0.918  & 0.139  & 24.5  & 0.510  \\
          & GTR~\cite{tu2023gaze}$\dagger$   & 0.962  & 0.068  & 14.2  & 0.597  \\
          & TransGOP~\cite{wang24transgop}$\dagger$ & 0.977  & 0.070  & 11.5  & 0.730 \\
          & TransGOP~\cite{wang24transgop} & 0.945  & 0.106  & 20.7  & - \\
\cmidrule{2-6}          & \cellcolor[rgb]{ .906,  .902,  .902}Ours$\dagger$ & \cellcolor[rgb]{ .906,  .902,  .902}0.978  & \cellcolor[rgb]{ .906,  .902,  .902}0.057  & \cellcolor[rgb]{ .906,  .902,  .902}10.2  & \cellcolor[rgb]{ .906,  .902,  .902}0.836  \\
          & \cellcolor[rgb]{ .906,  .902,  .902}Ours & \cellcolor[rgb]{ .906,  .902,  .902}0.938  & \cellcolor[rgb]{ .906,  .902,  .902}0.091  & \cellcolor[rgb]{ .906,  .902,  .902}18.9  & \cellcolor[rgb]{ .906,  .902,  .902}- \\
    \bottomrule
    \end{tabular}%
    }
\label{tab:ge_synth}
\end{minipage}
\quad 
\begin{minipage}[!b]{0.48\textwidth}
\makeatletter\def\@captype{table}
\centering
\caption{Gaze estimation results on GOO-Real dataset.  $\dagger$ indicates the evaluation method proposed by~\cite{tu2023gaze}.}
\resizebox{1 \linewidth}{!}{
\begin{tabular}{c|c|l|cccc}
    \toprule
    \multicolumn{3}{c|}{Method} & AUC    & Dist.$\downarrow$ & Ang.$\downarrow$  & mAP  \\
    \midrule
    \multirow{9}[6]{*}{\rotatebox{90}{No Pretrain}\quad} & \multirow{7}[4]{*}{\quad \rotatebox{90}{Non-Real}\quad} & GazeFollow~\cite{recasens2015they} & 0.850  & 0.220  & 44.4  & - \\
          &       & Lian~\cite{lian2018believe}  & 0.840  & 0.321  & 43.5  & - \\
          &       & VideoAttention~\cite{chong2020detecting} & 0.796  & 0.252  & 51.4  & - \\
          &       & GaTector~\cite{wang2022gatector} & 0.927  & 0.196  & 39.5  & - \\
          &       & Tonini~\cite{tonini2022multimodal} & 0.918  & 0.164  & -     & - \\
          &       & TransGOP~\cite{wang24transgop} & 0.947  & 0.097  & 16.7  & - \\
\cmidrule{3-7}          &       & Ours  & 0.943  & 0.078  & 12.9  & - \\
\cmidrule{2-7}          & \multirow{2}[2]{*}{\quad \rotatebox{90}{Real}\quad} & \cellcolor[rgb]{ .906,  .902,  .902}Ours$\dagger$ & \cellcolor[rgb]{ .906,  .902,  .902}0.971  & \cellcolor[rgb]{ .906,  .902,  .902}0.059  & \cellcolor[rgb]{ .906,  .902,  .902}9.8  & \cellcolor[rgb]{ .906,  .902,  .902}0.810 \\
          &       & \cellcolor[rgb]{ .906,  .902,  .902}Ours & \cellcolor[rgb]{ .906,  .902,  .902}0.944  & \cellcolor[rgb]{ .906,  .902,  .902}0.088  & \cellcolor[rgb]{ .906,  .902,  .902}14.7  & \cellcolor[rgb]{ .906,  .902,  .902}- \\
    \midrule
    \multirow{9}[6]{*}{\rotatebox{90}{Pretrain}\quad} & \multirow{7}[4]{*}{\quad \rotatebox{90}{Non-Real}\quad} & GazeFollow~\cite{recasens2015they} & 0.903  & 0.195  & 39.8  & - \\
          &       & Lian~\cite{lian2018believe}  & 0.890  & 0.168  & 32.6  & - \\
          &       & VideoAttention~\cite{chong2020detecting} & 0.889  & 0.150  & 29.1  & - \\
          &       & GaTector~\cite{wang2022gatector} & 0.940  & 0.087  & 14.8  & - \\
          &       & TransGOP~\cite{wang24transgop} & 0.957  & 0.081  & 14.7  & - \\
\cmidrule{3-7}          &       & Ours  & 0.963  & 0.073  & 12.4  & - \\
\cmidrule{2-7}          & \multirow{2}[2]{*}{\quad \rotatebox{90}{Real}\quad} & \cellcolor[rgb]{ .906,  .902,  .902}Ours$\dagger$ & \cellcolor[rgb]{ .906,  .902,  .902}0.978  & \cellcolor[rgb]{ .906,  .902,  .902}0.051  & \cellcolor[rgb]{ .906,  .902,  .902}8.8  & \cellcolor[rgb]{ .906,  .902,  .902}0.824  \\
          &       & \cellcolor[rgb]{ .906,  .902,  .902}Ours & \cellcolor[rgb]{ .906,  .902,  .902}0.949  & \cellcolor[rgb]{ .906,  .902,  .902}0.082  & \cellcolor[rgb]{ .906,  .902,  .902}13.8  & \cellcolor[rgb]{ .906,  .902,  .902}- \\
    \bottomrule
    \end{tabular}%

    }
\label{tab:ge_real}
\end{minipage}
\end{figure}

\begin{table*}[!t]
  \centering
  \caption{Ablation study about each component on GOO-Real}
  \resizebox{0.9\linewidth}{!}{
    \begin{tabular}{ccccc|ccc|ccc|ccc}
    \toprule
    \multirow{2}[3]{*}{PS} & \multirow{2}[3]{*}{RoI-Rec.} & \multirow{2}[3]{*}{DAF} & \multirow{2}[3]{*}{FIM} & \multirow{2}[3]{*}{EAL} & \multicolumn{3}{c|}{Gaze Object Segmentation} & \multicolumn{3}{c|}{Instance Segmentation} & \multicolumn{3}{c}{Gaze Estimation} \\
\cmidrule{6-14}          &       &       &       &       & mSoC & mSoC$_{50}$ & mSoC$_{75}$ & AP   & \quad  AP$_{50}$ & AP$_{75}$ & AUC    & Dist.$\downarrow$ & Ang.$\downarrow$ \\
\midrule
          &       &       &       &       &   79.00    &   98.48    &   92.50    & 74.24  & 98.00  & 87.62  & 0.854  & 0.304  & 54.2  \\
    $\checkmark$ &       &       &       &       &    84.50   &   98.75    &   96.56    & 80.53  & 98.68  & 93.67  & 0.839  & 0.334  & 59.8  \\
     $\checkmark$ &  $\checkmark$ &       &       &       &   84.63    &    98.67   &    96.72   & 81.70  & 98.83  & 94.66  & 0.910  & 0.114  & 20.5  \\
     $\checkmark$ &  $\checkmark$ &  $\checkmark$ &       &       &    85.44   &   98.92    &    96.96   & 81.58  & 98.76  & 94.29  & 0.919  & 0.095  & 16.8  \\
     $\checkmark$ &  $\checkmark$ &  $\checkmark$ &  $\checkmark$ &       &    85.89   &    98.80   &    97.14   & 82.45  & 98.92  & 94.82  & 0.932  & 0.089  & 14.9  \\
    \rowcolor[rgb]{ .906,  .902,  .902}  $\checkmark$ &  $\checkmark$ &  $\checkmark$ &  $\checkmark$ &  $\checkmark$ &85.94    &    98.99   &    97.20& 82.86  & 99.02  & 95.57  & 0.944  & 0.088  & 14.7  \\
    \bottomrule
    \end{tabular}%
    }
  \label{tab:ablation}%
\end{table*}%

\textbf{Ablation study about the effect of each component.}
In Table~\ref{tab:ablation}, we conduct various ablation studies on the GOO-Real dataset. We first establish a baseline using MaskDINO~\cite{li2023mask} and a gaze regression branch similar to TransGOP~\cite{wang24transgop}, and then gradually add the proposed modules and demonstrate their contribution to overall model performance.
\textbf{(i) Pixel shuffle (PS).} 
After adding the pixel shuffle module, instance segmentation performance improved by 6.29\% AP (80.53\% vs 74.24\%) compared to baseline, indicating that large feature maps significantly enhance detailed features in dense prediction tasks. 
\textbf{(ii) RoI Reconstruction (RoI-Rec.).} The RoI reconstruction module reduces angle error by 39.3 and L2 distance by 0.22, demonstrating the importance of head-related features for gaze direction perception and the ability of the module to learn accurate gaze features in real-world settings.
\textbf{(iii) Dual attention fusion (DAF).} Introducing the dual attention fusion improves gaze heatmap quality (AUC: 0.919 vs 0.910), demonstrating that perceiving gaze object spatial location beforehand produces more accurate heatmaps and eases regression.
\textbf{(iv) Feature interaction module (FIM).} 
The feature interaction module improves gaze estimation, reducing angle error by 1.9 and L2 distance by 0.006. Gaze object segmentation achieves an 85.89 mSoC, showing that pixel-level object information enhances feature interaction and human-object gaze relationship modeling.
\textbf{(v) Energy aggregation loss (EAL).} Adding energy aggregation loss improves the performance of all three sub-tasks, showing that the gaze object mask effectively guides the regression of the gaze heatmap towards the gaze object.

\begin{table*}[!t]
  \centering
  \caption{Analysis about dual attention fusion module.}
  \resizebox{0.9\linewidth}{!}{
    \begin{tabular}{cc|ccc|ccc|ccc}
    \toprule
    \multirow{2}[3]{*}{Original Attention} & \multirow{2}[3]{*}{Gaze Cone} & \multicolumn{3}{c|}{Gaze Object Segmentation} & \multicolumn{3}{c|}{Instance Segmentation} & \multicolumn{3}{c}{Gaze Estimation} \\
\cmidrule{3-11}          &       & mSoC   & mSoC$_{50}$ & mSoC$_{75}$ & AP    & AP$_{50}$ & AP$_{75}$ & AUC    & Dist.$\downarrow$ & Ang.$\downarrow$ \\
\midrule
    $\checkmark$ &    & 85.51   &    98.86   &    97.17  &   82.08    &   98.87    &    94.78   &    0.938   &   0.089    & 15.6 \\
          & $\checkmark$ &    84.13   &    98.78   &    96.35   &    81.30   &   98.78   &    93.69   &   0.926    &   0.097    & 16.3 \\
    \rowcolor[rgb]{ .906,  .902,  .902} $\checkmark$ & $\checkmark$ &    85.94   &    98.99   &   97.20    & 82.75  & 99.01  & 95.17  & 0.944  & 0.088  & 14.7  \\
    \bottomrule
    \end{tabular}%
    }
  \label{tab:chor}%
\end{table*}%

\begin{table*}[!t]
  \centering
  \caption{Analysis about feature interaction module}
  \resizebox{0.9\linewidth}{!}{
    \begin{tabular}{ccc|ccc|ccc|ccc}
    \toprule
    \multirow{2}[3]{*}{\tabincell{c}{Encoder \\ Query}} & \multirow{2}[3]{*}{\tabincell{c}{Decoder \\ Query}} & \multirow{2}[3]{*}{\tabincell{c}{Mask \\ Embeding}} & \multicolumn{3}{c|}{Gaze Object Segmentation} & \multicolumn{3}{c|}{Instance Segmentation} & \multicolumn{3}{c}{Gaze Estimation} \\
\cmidrule{4-12}          &       &       & mSoC   & mSoC$_{50}$ & mSoC$_{75}$ & AP    & AP$_{50}$ & AP$_{75}$ & AUC    & Dist.$\downarrow$ & Ang.$\downarrow$ \\
\midrule
    $\checkmark$ &       &       &    85.79   &    98.38   &    96.44   &  81.28     &   98.87    &   94.87    &    0.928   &   0.095    &  16.0 \\
          & $\checkmark$ &       &    85.20   &    98.21   &    97.01   & 81.65  & 98.86  & 94.65  & 0.949  & 0.093  & 15.1  \\
    \rowcolor[rgb]{ .906,  .902,  .902}       &       & $\checkmark$ & 85.94    &    98.99   &    97.20& 82.86  & 99.02  & 95.57  & 0.944  & 0.088  & 14.7  \\
    \bottomrule
    \end{tabular}%
    }
  \label{tab:ca}%
\end{table*}%

\noindent \textbf{Analysis about dual attention fusion module.}
Table~\ref{tab:chor} analyzed the effect of different spatial perception methods. Compared to using original attention $\mathbf{M}_{\rm ori}$ or gaze cone $\mathbf{M}_{\rm cone}$ alone, our dual attention fusion performs better. This indicates that combining spatial search along the gaze direction with human head-based scene saliency is more effective. Fig.~\ref{fig:chor_module}(b) shows that element-wise fusion of $\mathbf{M}_{\rm ori}$ and $\mathbf{M}_{\rm cone}$ creates an initial human-object spatial connection.

\noindent \textbf{Analysis about feature interaction module.}
In Table~\ref{tab:ca}, we evaluate the efficacy of different object information for gaze regression. Injecting the global encoder query of MaskDINO~\cite{li2023mask} results in 85.79\% mSoC for gaze object segmentation but poor gaze estimation performance. Refining the decoder query with mask supervision reduces the angle error by 0.9, indicating better object representation. Using mask embedding with mask features during gaze regression achieves optimal performance, demonstrating that semantically clear mask features enhance gaze object positioning and the human-object gaze relationship.

\begin{figure*}[!t]
\centering
\includegraphics[width=1\linewidth]{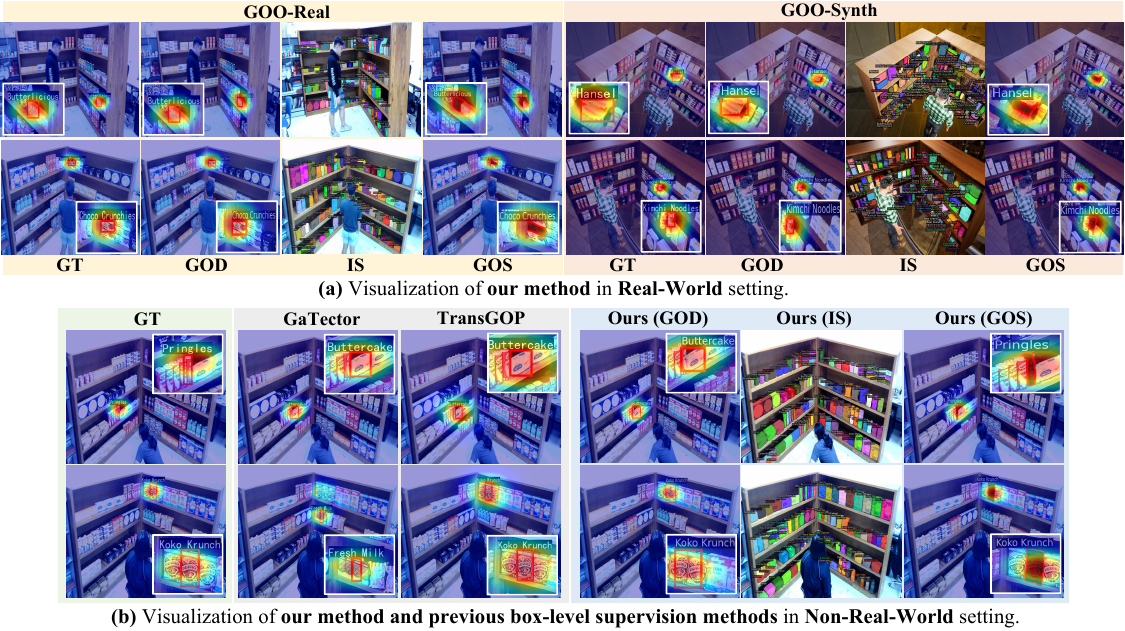}
\caption{Qualitative comparison of ground truth gaze object (GT), gaze object detection (GOD), instance segmentation
(IS), gaze object segmentation (GOS)}
\label{fig:vis}
\end{figure*}

\subsection{Qualitative Results.}
Fig.~\ref{fig:vis}(a) presents qualitative results in the real-world setting. In object-dense scenes, our method accurately predicts the gaze object mask without relying on head-related cues. In Fig.~\ref{fig:vis}(b), compared to previous methods, our method accurately locates adjacent objects using a pixel-level mask, effectively alleviating semantic ambiguity. See \textbf{supplementary material} for more visualization.

\section{Conclusion}
In this paper, we introduce the gaze object segmentation task, which aims to infer pixel-level masks for objects captured by human eyesight, and establish a framework based on the pixel-level supervision from VFM to solve this task while alleviating the semantic ambiguity existing in previous box-level supervision methods. In our method, we choose SAM to generate pixel-level
supervision information for the model training. The RoI reconstruction module is proposed to reconstruct head features directly from holistic features using the head box provided by the unified Transformer, which improves the inference efficiency and flexibility of our model in the real world. To boost the representation of the human-object gaze relationship, we propose a space-to-object gaze regression strategy, which first uses the dual attention fusion module to establish a human-object spatial connection, and then a semantically clear mask feature is injected into the gaze regression to strengthen the human-object gaze relationship modeling. Extensive experiments on GOO-Real and GOO-Synth demonstrate the effectiveness of our method in various settings.

\noindent \textbf{Limitation:} The proposed method requires box supervision as prompts for the SAM when generating masks, which limits its application in a wider range of scenarios without strong location priors, more flexible approaches for generating masks via the latest VFM models need to be explored in future work. 

\section*{Acknowledgements}
This work is supported by the National Natural Science Foundation of China. (No. 82272020). 

\noindent Bin Fan is sponsored by China National Postdoctoral Program for Innovative Talents (No. BX20230013).

\bibliographystyle{splncs04}
\bibliography{main}

\clearpage
\appendix

\begin{center}
\textbf{\large Supplementary Materials of Boosting Gaze Object Prediction via Pixel-level Supervision from Vision Foundation Model}
\end{center}

\section{Gaze object detection compared to GTR~\cite{tu2023gaze}}
Recently, Tu~\etal proposed GTR~\cite{tu2023gaze} to jointly predict gaze location and gaze object but only did limited validation for gaze object detection on the GOO-Real dataset. In GTR~\cite{tu2023gaze}, the AP of object detection is used to evaluate gaze object detection, and mean average precision (mAP) is used to evaluate the detection accuracy of multiple indicators. In the process of calculating mAP, a predicted gaze instance detection is considered a true positive if and only if it is highly accurate (\ie, the IoU between the predicted human head box and ground truth is greater than 0.5 while the L2 distance is less than 0.15 and confidence for gaze object is larger than 0.75), and only calculate AUC and L2 distance for those predicted instances considered as true positives, thereby ignoring relatively poor detection instances, resulting in an incomplete evaluation. For a fair comparison, we also report the performance of our model using the same metric in Table~\ref{tab:compare_gtr}. Our model outperforms GTR by 19.06 (77.32\% \vs 58.26\%) and 18.08 (75.20\% \vs 57.12\%) in real-world and non-real-world settings respectively. At the same time, mAP also increased by 0.183 (0.810 \vs 0.627) in real-world setting compared to GTR, which indicates that compared to directly using the Transformer to globally model the relationship between head and gazed location, our special attention to the human head produced by RoI reconstruction module and space-to-object gaze regression approach can more effectively perceive the gaze direction and search gaze object in the corresponding area, achieving more accurate detection results.

\begin{table}[htbp]
  \centering
  \caption{Gaze object detection results on GOO-Real dataset. For a fair comparison, we use the same evaluation metrics as GTR~\cite{tu2023gaze}. $\ast$ represents the results provided by GTR~\cite{tu2023gaze}.}
  \vspace{-10pt}
  \resizebox{0.7\linewidth}{!}{
    \begin{tabular}{l|ccc|ccc|c}
    \toprule
    \multicolumn{1}{c|}{\multirow{2}[4]{*}{Method}} & \multicolumn{3}{c|}{Non-Real} & \multicolumn{4}{c}{Real} \\
\cmidrule{2-8}          & AP    & AP$_{50}$ & AP$_{75}$ & AP    & AP$_{50}$ & AP$_{75}$ & mAP \\
    \midrule
    GazeFollow~\cite{recasens2015they}$\ast$ & 35.5  & 58.9  & 26.7  & 31.6  & 54.3  & 24.1  & 0.291  \\
    Lian~\cite{lian2018believe}$\ast$  & 35.7  & 62.8  & 29.4  & 32.1  & 58.4  & 27.2  & 0.297  \\
    VideoAttention~\cite{chong2020detecting}$\ast$ & 36.8  & 61.5  & 30.8  & 32.4  & 56.7  & 27.9  & 0.314  \\
    GaTector~\cite{wang2022gatector}$\ast$ & 52.25  & 91.92  & 55.34  & 48.70  & 86.40  & 52.10  & 0.437  \\
    GTR~\cite{tu2023gaze}$\ast$ & 58.26  & 97.31  & 58.42  & 57.12  & 96.47  & 58.90  & 0.627  \\
    \midrule
    \rowcolor[rgb]{ .906,  .902,  .902} Ours  & \textbf{77.32}  & \textbf{98.65}  & \textbf{93.91}  & \textbf{75.20}  & \textbf{98.94}  & \textbf{93.20}  & \textbf{0.810}  \\
    \bottomrule
    \end{tabular} }
  \label{tab:compare_gtr}%
\end{table}%

\section{Energy aggregation loss}
In this paper, following TransGOP~\cite{wang24transgop}, we use an energy aggregation loss to further guide the gaze heatmap to converge toward the location of the gazed object. Since we perform gaze object segmentation, we replace the bounding box with a mask for calculation to more accurately measure the error, which can be formulated as:
\begin{equation}
    \mathcal{L}_{\rm eng} = {1} - \frac{1}{S_{mask}} \sum_{(i, j) \in \text{mask}} \textbf{M}_{i,j}
    \label{eq:eng_loss}
\end{equation}
where $S_{mask}$ represents the area of the gaze object mask and $\textbf{M}$ represents the gaze heatmap.

\section{Experiments Settings}
\noindent \textbf{Datasets:} We conduct a comprehensive evaluation of our model on the GOO-Synth and GOO-Real datasets. The GOO-Synth dataset contains 192,000 synthetic images, 172,000 images for training, and 2,0000 images for testing. Each image contains approximately 80 objects distributed in 24 categories and each object has accurate bounding box annotation. In addition, the bounding box of the human head in the image and its corresponding gaze point and gaze object bounding box are also annotated. The GOO-Real contains 9,552 real-world images and has the same annotation information as the GOO-Synth dataset. Both GOO-Synth and GOO-Real image scenes contain a large number of small and dense objects and the subjects always face away from the camera, which is very challenging for gaze estimation, gaze object segmentation, and detection. 

\noindent \textbf{Implementation Details:} The backbone $\mathcal{B}$ is a ResNet-50~\cite{he2016deep} pre-trained on ImageNet~\cite{krizhevsky2012imagenet}. We use MaskDINO~\cite{li2023mask} as the detection transformer to extract pixel-level mask features and detect the bounding boxes and instance masks of candidate gaze objects. We train all our components with AdamW~\cite{loshchilov2017decoupled} optimizer and a learning rate of 0.0001 for 75 epochs with a batch size of 2. In the main text Eq. (4), we set $\alpha$, $\beta$, and $\gamma$ as 10, 1000, and 1 respectively. We set the pixel shuffle scale factor $\eta$ as 2, the input size as $224\times224$, and the predicted heatmap size as $64\times 64$. All experiments are implemented based on the PyTorch and one GeForce RTX 3090Ti GPU.

\noindent \textbf{Metrics:} Our mask-based mSoC calculation method is defined as follows:
\begin{equation}
    mSoC_{mask} = \min{(\frac{p_{mask}}{a}, \frac{g_{mask}}{a})} \times \frac{{p_{mask}} \cup {g_{mask}}}{a}.
    \label{mSoC}
\end{equation}
where $p_{mask}$ and $g_{mask}$ represent the areas of the prediction mask and ground truth mask respectively, and $a$ represent the area of the minimum enclosing rectangle of $p_{mask}$ and $g_{mask}$. It is worth noting that compared to using bounding box calculations, mask-based calculation methods are more restrictive because pixel-level positioning is more accurate.

\begin{table*}[!t]
  \centering
  \caption{Analysis about RoI reconstruction module}
  \resizebox{0.9\linewidth}{!}{
    \begin{tabular}{cc|ccc|ccc|ccc}
    \toprule
    \multirow{2}[3]{*}{RoI-Pooling} & \multirow{2}[3]{*}{RoI-Align} & \multicolumn{3}{c|}{Gaze Object Segmentation} & \multicolumn{3}{c|}{Instance Segmentation} & \multicolumn{3}{c}{Gaze Estimation} \\
\cmidrule{3-11}          &       & mSoC  & mSoC$_{50}$ & mSoC$_{75}$ & AP    & AP$_{50}$ & AP$_{75}$ & AUC$\uparrow$   & Dist.$\downarrow$ & Ang.$\downarrow$ \\
\midrule
    $\checkmark$ &       &    84.21   &    98.90   &    96.57   & 81.36  & 98.84  & 94.52  & 0.942  & 0.092  & 15.1  \\
    \rowcolor[rgb]{ .906,  .902,  .902}       & $\checkmark$ &    85.94   &    98.99   &    97.20   & 82.86  & 99.02  & 95.57  & 0.944  & 0.088  & 14.7  \\
    \bottomrule
    \end{tabular}%
    }
  \label{tab:roi-rec}%
\end{table*}%

\begin{figure}[!t]
\centering
\includegraphics[width=1\linewidth]{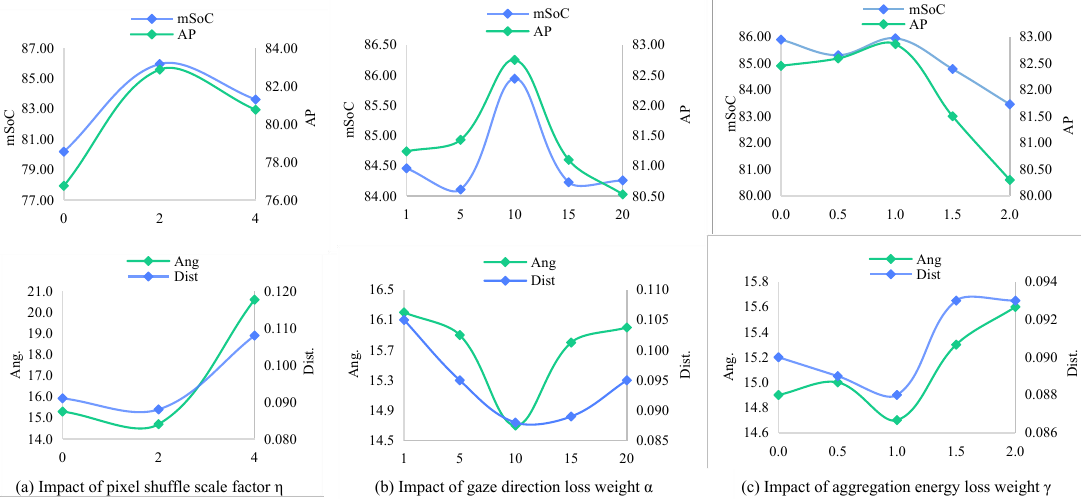}
\caption{Impact of hyper-parameter $\eta$, $\alpha$, and $\gamma$.}
\label{fig:hyper}
\end{figure}

\section{Analysis about RoI reconstruction module}
In this paper, we reconstruct head region features based on overall features to improve the efficiency and flexibility of the model in the real world. We explore the effectiveness of different reconstruction methods. As shown in Table~\ref{tab:roi-rec}, we first use RoI pooling to crop head features directly from the original features, which obtains comparable results. Based on this, we further use RoI align to reconstruct the head feature map, which uses more detailed bilinear interpolation to restore feature resolution and achieve more detailed feature representation. 

\section{Analysis about hyper-parameters $\eta$, $\alpha$, and $\gamma$}
To explore the optimal performance of the model and sensitivity, we adjusted the hyper-parameters in the model on the GOO-Real dataset. We first adjusted the pixel shuffle scale factor $\eta$. As shown in Fig~\ref{fig:hyper}(a), the gaze object prediction and gaze estimation performance achieve best results when $\eta$ = 2. When the feature map continues to be enlarged, the performance decreases instead, which shows that using smaller inputs to generate too large feature maps will introduce noise in the learning process, resulting in unclear representations. We explore the sensitivity of gaze direction loss in Fig~\ref{fig:hyper}(b). The model achieves optimal performance when $\alpha$ equals 10. Fig~\ref{fig:hyper}(c) shows the impact of aggregation energy loss weight $\gamma$. When $\gamma$ equals 1, the model performance is optimal. When $\gamma$ continues to increase, the model performance will affect the performance of multi-task learning, resulting in performance degradation.

\section{More visualization}
\begin{figure}[!t]
\centering
\includegraphics[width=1\linewidth]{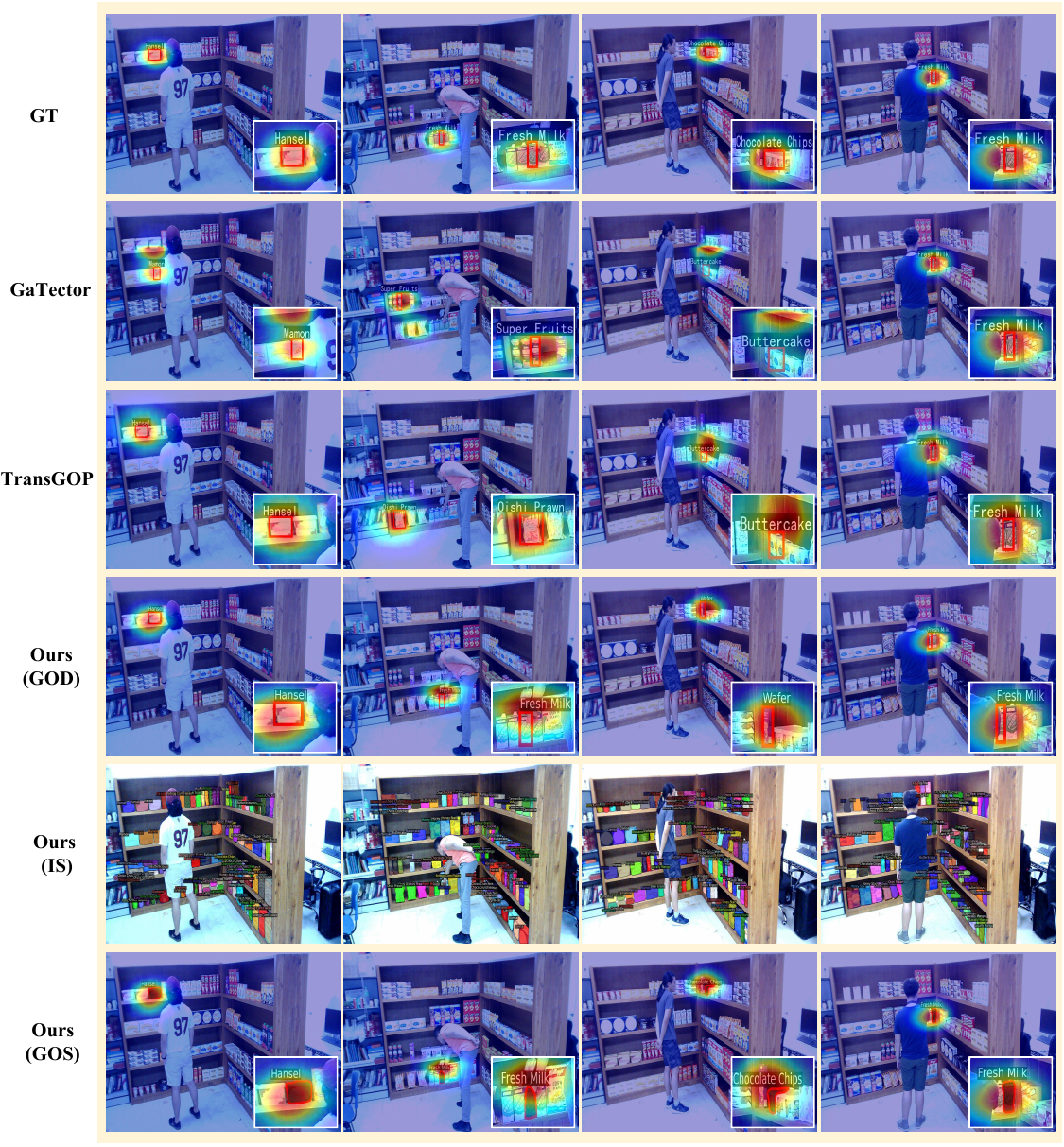}
\caption{Visualization of \textbf{our real-world model} and \textbf{previous non-real world model} on GOO-Real dataset. We show the results of ground truth gaze object (GT), gaze object detection (GOD), instance segmentation (IS), gaze object segmentation (GOS), and previous methods.}
\label{fig:more_vis_compare}
\end{figure}

\begin{figure}[!t]
\centering
\includegraphics[width=1\linewidth]{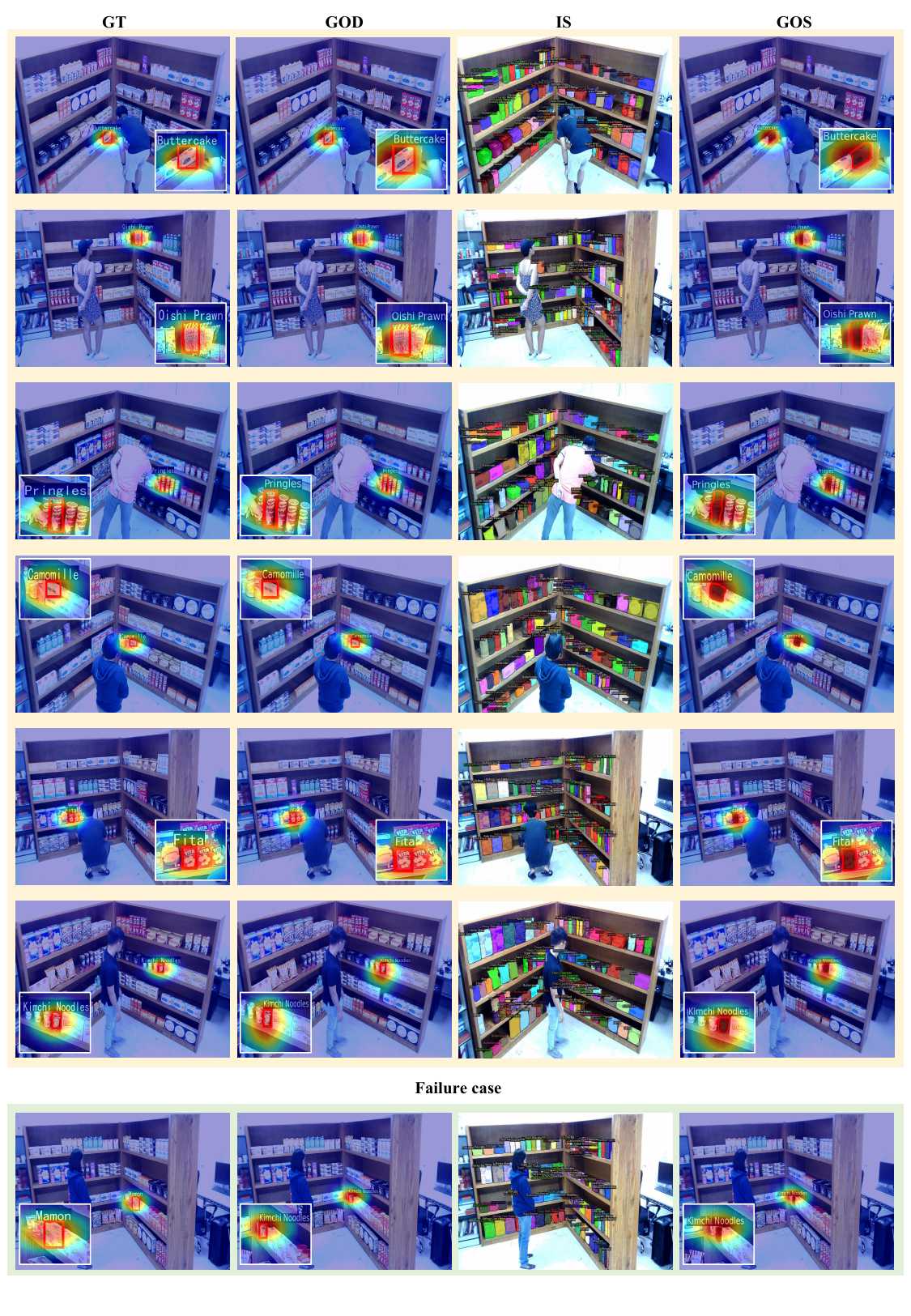}
\caption{Visualization of our method in Real-World setting on GOO-Real dataset. We show the results of ground truth gaze object (GT), gaze object detection (GOD), instance segmentation (IS), and gaze object segmentation (GOS). The last row shows the failure case.}
\label{fig:more_vis_real}
\end{figure}

\begin{figure}[!t]
\centering
\includegraphics[width=1\linewidth]{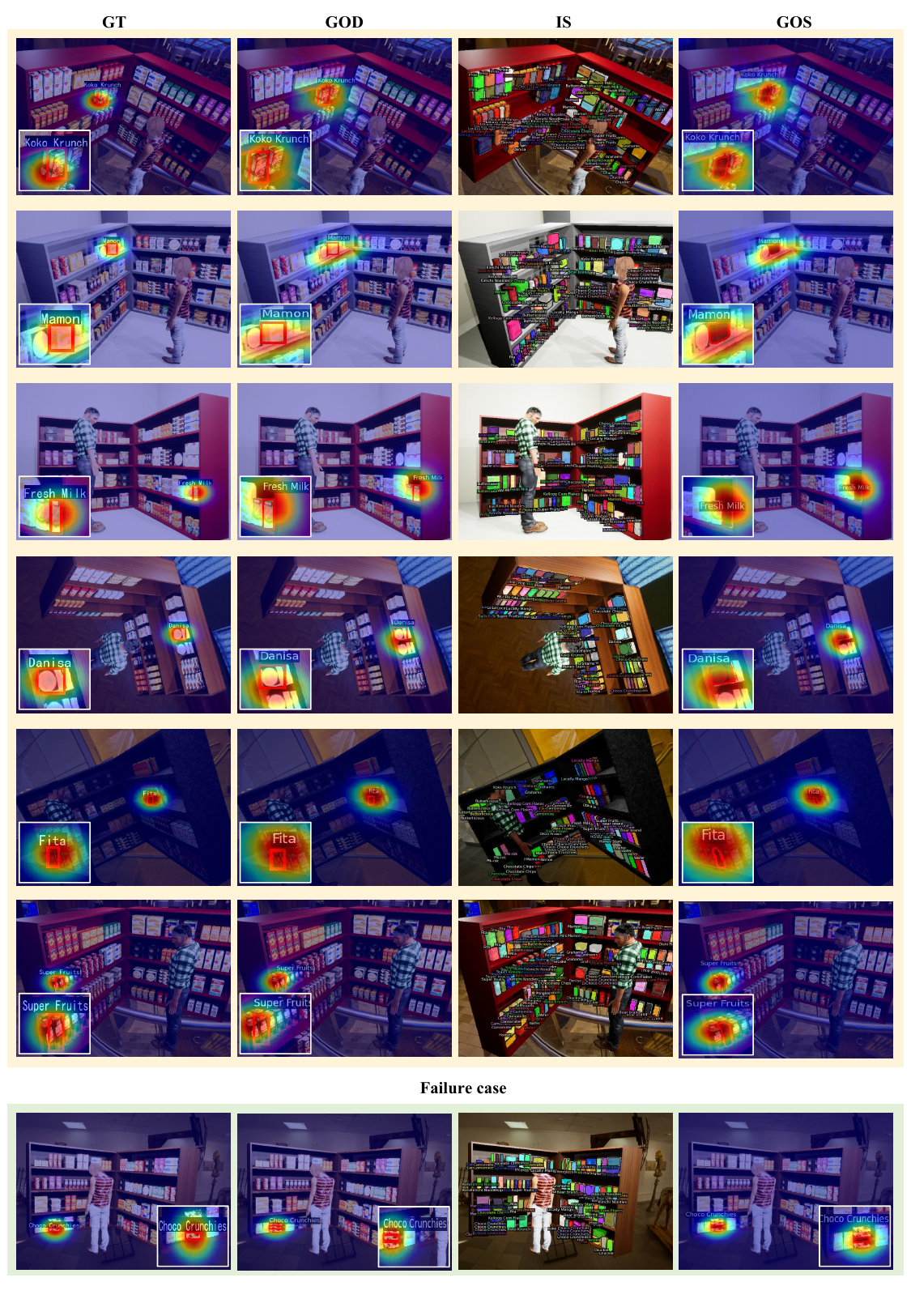}
\caption{Visualization of our method in Real-World setting on GOO-Synth dataset. We show the results of ground truth gaze object (GT), gaze object detection (GOD), instance segmentation (IS), and gaze object segmentation (GOS). The last row shows the failure case.}
\label{fig:more_vis_synth}
\end{figure}

Fig~\ref{fig:more_vis_compare} shows the visualization results of our method and previous methods. GaTector and TransGOP regress inaccurate gaze heatmaps due to the lack of spatial perception and detailed features understanding of object locations. In contrast, our model achieves more accurate gaze object positioning through pixel-level prediction and space-to-object gaze regression method, thereby alleviating semantic ambiguity while regressing a more accurate gaze heatmap. Fig~\ref{fig:more_vis_real} and Fig~\ref{fig:more_vis_synth} show the visualization results of our method on GOO-Real and GOO-Synth datasets respectively. Our method can regress accurate gaze heatmap for gaze objects and achieve pixel-level prediction. Our model still achieves relatively accurate predictions even when the person's back is turned to the camera. We also show failure cases, mainly because the person is relatively far away from the object, making it difficult to model the pixel-level gaze relationship, which indicates optimization directions for future works.

\end{document}